# Clinically Interpretable Biological Age Estimation Using CatBoost and SHAP with Continuous and Categorical Features from NHANES

Running Title: Clinically Interpretable Biological Age Estimation

**Yanni Sun, MS[1]; Zheng Qu, PhD[1]; Kevin Ren, PhD[1]; Zhanglin Lin Cui, PhD[1]***

[1]Northern Center for Learning and Innovation, Middletown, New York, USA

*Corresponding author, zlcui@mt.feitian.edu

## Abstract

Biological age (BA) is intended to capture inter-individual variation in physiological aging beyond chronological age (CA). Although machine-learning models have advanced BA estimation, many rely primarily on continuous laboratory biomarkers, and SHapley Additive exPlanations (SHAP) dependence plots have generally been interpreted visually. We developed a clinically interpretable machine-learning framework for BA estimation using heterogeneous clinical data from the National Health and Nutrition Examination Survey (NHANES). Adults aged ≥18 years from NHANES 2011–March 2020 were included. A CatBoost regression model used 44 demographic, laboratory, examination, and lifestyle predictors to estimate CA, with model-estimated age interpreted as BA. SHAP provided population-, feature-, and individual-level interpretation. Building on previous visual interpretation approaches, we applied a prespecified automated procedure directly to SHAP dependence plots to calculate feature-value ranges associated with the lowest and highest BA contributions, corresponding BA adjustment magnitudes, and SHAP zero-crossing points. The analytical cohort included 24,983 adults. Model-estimated BA was strongly correlated with CA in the held-out testing dataset (Pearson $r = 0.798$, $P < 0.001$). Population-level SHAP analysis identified blood urea nitrogen, serum glucose, vigorous recreational activity, red cell distribution width (RDW), and alanine aminotransferase among the leading predictors. Three categorical predictors ranked among the 25 most influential variables. Automated quantitative SHAP analysis characterized predictor-specific feature ranges, BA adjustment magnitudes, and SHAP zero-crossing points, while waterfall plots provided participant-specific explanations. This clinically interpretable framework integrates heterogeneous clinical data with multi-level model interpretation and automated quantitative SHAP dependence analysis. By extending visual SHAP interpretation to prespecified, reproducible quantitative calculation, the framework enhances transparency and provides a methodological foundation for biological aging research and future precision-health applications.

## 1. Introduction

Population aging has emerged as one of the greatest public health challenges of the twenty-first century, contributing substantially to the global burden of chronic diseases, disability, healthcare utilization, and socioeconomic costs worldwide (Kennedy et al., 2014; Levine & Crimmins, 2018). Aging is a progressive biological process characterized by the accumulation of molecular and cellular damage that gradually impairs physiological function and increases vulnerability to age-related diseases (Kennedy et al., 2014; López-Otín et al., 2013). As life expectancy continues to increase globally, extending healthspan, rather than lifespan alone, has become a major objective of geroscience and precision medicine (Kennedy et al., 2014; Levine & Crimmins, 2018). Achieving this goal requires reliable methods for quantifying the biological aging process, identifying individuals at increased risk of adverse health outcomes, and evaluating interventions designed to promote healthy aging (Jia et al., 2017; Levine et al., 2018).

Chronological age (CA), defined as the time elapsed since birth, remains the most widely used measure of aging because it is objective, universally available, and strongly associated with many physiological changes (Jia et al., 2017). However, considerable heterogeneity exists among individuals of the same CA because biological aging is influenced by genetic predisposition, environmental exposures, lifestyle behaviors, and chronic disease (Levine & Crimmins, 2018; Lu, 2024). Consequently, CA alone cannot adequately characterize physiological aging or distinguish individuals who are aging more rapidly or more slowly than expected (Jia et al., 2017; Kennedy et al., 2014).

To better capture these individual differences, the concept of biological age (BA) has been proposed as an estimate of an individual's physiological status relative to that expected for a given CA (Jia et al., 2017; Levine & Crimmins, 2018). Unlike CA, BA integrates the cumulative effects of genetics, lifestyle, environmental exposures, and disease on physiological function (Levine et al., 2018; Lu, 2024). BA measures have been associated with mortality, frailty, functional decline, and multiple age-related diseases, highlighting their potential value as biomarkers of healthy aging (Levine et al., 2018; Liu et al., 2018). Consequently, accurate estimation of BA has become an increasingly important objective in aging research and preventive medicine (Bernard et al., 2023; Qiu et al., 2023).

Recent advances in machine learning have transformed BA estimation by enabling nonlinear modeling of complex interactions among large numbers of physiological, biochemical, and clinical variables (Bae et al., 2021; Mamoshina et al., 2018). Machine-learning algorithms such as Random Forest, Support Vector Regression, Extreme Gradient Boosting (XGBoost), and Deep Neural Networks have demonstrated strong predictive performance across BA studies and can capture nonlinear relationships that may be difficult to represent with conventional statistical models (Bae et al., 2021; Mamoshina et al., 2018). More recently, interpretable tree-based models have further advanced the field by combining accurate prediction with biologically meaningful explanations of model behavior. In particular, Wood et al. developed an interpretable XGBoost model that integrated SHapley Additive exPlanations (SHAP) to quantify feature importance and visualize participant-specific BA predictions, providing one of the first clinically interpretable machine learning frameworks for BA estimation (Wood et al., 2019).

As machine learning models become increasingly sophisticated, model interpretability has emerged as a critical requirement for clinical translation (Bernard et al., 2023; Prosz et al., 2024).

High predictive accuracy alone is insufficient if clinicians cannot understand how individual variables influence model predictions or identify the factors contributing to accelerated biological aging (Bernard et al., 2023). Explainable artificial intelligence (XAI) addresses this challenge by improving model transparency while preserving predictive performance (Qiu et al., 2023). Among currently available XAI techniques, SHAP has become one of the most widely adopted methods for interpreting tree-based machine learning models because it provides theoretically consistent feature attribution at both the population and individual levels (Doumard et al., 2023; Lundberg & Lee, 2017). Recent frameworks, including ExplaiNAble BioLogical Age (ENABL Age) and personalized physiological age models, further demonstrate that integrating machine learning with XAI can generate clinically meaningful interpretations of biological aging while facilitating individualized risk assessment (Bernard et al., 2023; Qiu et al., 2023).

Despite these advances, an important methodological gap remains. Most published BA models have been developed primarily from continuous laboratory biomarkers, whereas real-world clinical datasets routinely combine laboratory measurements with demographic characteristics, lifestyle behaviors, and other categorical clinical variables (Bae et al., 2021; Mamoshina et al., 2018; Wood et al., 2019). Efficiently integrating these heterogeneous data types while maintaining model interpretability remains challenging because conventional machine learning algorithms often require extensive feature engineering or categorical encoding before model development (Prokhorenkova et al., 2018). In addition, the landmark study by Wood et al. (2019) demonstrated the value of SHAP dependence plots for interpreting BA models by visually estimating informative biomarker ranges, BA adjustment magnitudes, and approximate transition points. Subsequent work has also demonstrated the value of quantitative SHAP-based metrics for personalized physiological age interpretation (Bernard et al., 2023). Building on these foundations, automated calculation directly from SHAP dependence plots of predictor-specific extreme-contribution ranges, BA adjustment magnitudes, and SHAP zero-crossing points offers an opportunity to improve reproducibility and quantitative interpretability.

Categorical Boosting (CatBoost) is a gradient boosting algorithm specifically designed to process heterogeneous datasets containing both continuous and categorical variables through ordered target statistics while minimizing prediction bias and overfitting (Prokhorenkova et al., 2018). Its native handling of categorical predictors makes it particularly attractive for BA estimation, where demographic, behavioral, and lifestyle variables may provide complementary information beyond conventional laboratory biomarkers. Leveraging these methodological advantages may facilitate the development of clinically interpretable BA estimation models that more closely reflect the heterogeneous nature of human aging while preserving transparent explanations of both population-level patterns and individualized predictions.

Therefore, the present study aimed to develop a clinically interpretable machine-learning framework for BA estimation by integrating CatBoost with SHAP using continuous and categorical features derived from the National Health and Nutrition Examination Survey (NHANES). The framework was designed around three complementary methodological advances: integration of heterogeneous demographic, laboratory, examination, and lifestyle variables; multi-level interpretation spanning population-, feature-, and individual-level explanations; and automated quantitative interpretation of SHAP dependence plots. Building on the visual interpretation approach established by Wood et al. (2019), we algorithmically calculated predictor-value ranges associated with the lowest and highest SHAP values, the corresponding BA adjustment magnitudes, and predictor-specific SHAP zero-crossing points. Collectively, these analyses extend interpretable BA estimation from predictor ranking and visualization toward reproducible quantitative and individualized interpretation.

## 2. Materials and Methods

This study developed and evaluated a clinically interpretable machine-learning framework for biological age (BA) estimation using heterogeneous clinical data from the National Health and Nutrition Examination Survey (NHANES). The methodological workflow comprised study population selection, construction of predictor and target variables, CatBoost model development, evaluation in a held-out testing dataset, and model interpretation at the population, feature, and individual levels using SHAP. In this study, clinically interpretable refers to the use of familiar demographic, laboratory, examination, and lifestyle variables together with transparent explanations at the population, feature, and individual levels; it does not imply established clinical utility or validated decision thresholds.

### 2.1 Study Design and Data Source

This retrospective cross-sectional study utilized publicly available data from NHANES collected between 2011 and March 2020. NHANES is an ongoing nationally representative health survey conducted by the National Center for Health Statistics (NCHS), Centers for Disease Control and Prevention (CDC), using a complex multistage probability sampling design to assess the health and nutritional status of the noninstitutionalized U.S. population. The survey combines standardized household interviews with physical examinations and laboratory testing performed in mobile examination centers, providing comprehensive demographic, clinical, behavioral, and laboratory information for epidemiologic research (National Center for Health Statistics, 2018, 2022; Wood et al., 2019).

All NHANES participants provided written informed consent, and the survey protocol was approved by the NCHS Research Ethics Review Board. Because NHANES data are publicly available and fully de-identified, this secondary analysis did not require additional institutional review board approval.

The present study followed the methodological framework of first-generation BA models by developing a supervised machine learning model to estimate chronological age (CA) from multidimensional clinical variables, with the predicted age interpreted as BA, consistent with previous BA studies (Bae et al., 2021; Wood et al., 2019). Accordingly, the resulting BA estimate should be understood as a model-derived age phenotype reflecting age-related patterns in the included predictors rather than a direct measurement of the underlying biological aging process.

### 2.2 Study Population

NHANES datasets from 2011 through March 2020 were downloaded and merged using the unique participant identifier (SEQN). Demographic, laboratory, examination, and questionnaire datasets were linked to generate a unified analytical dataset.

The merged dataset initially contained 45,462 participants. Participants younger than 18 years were excluded to restrict the analysis to adults. Consistent with previous NHANES BA studies, participants with missing values in more than 17 variables were also excluded to ensure adequate information for model development (Wood et al., 2019). Duplicate records were removed during data integration.

After applying the eligibility criteria, the final analytical cohort consisted of 24,983 adults. All retained pre-specified 44 predictor variables had fewer than 5% missing observations, and no additional variable-specific exclusion was required. No missing values were manually imputed prior to CatBoost modeling. The missing values were handled natively and automatically by CatBoost. A participant flow diagram is presented in Figure 1.

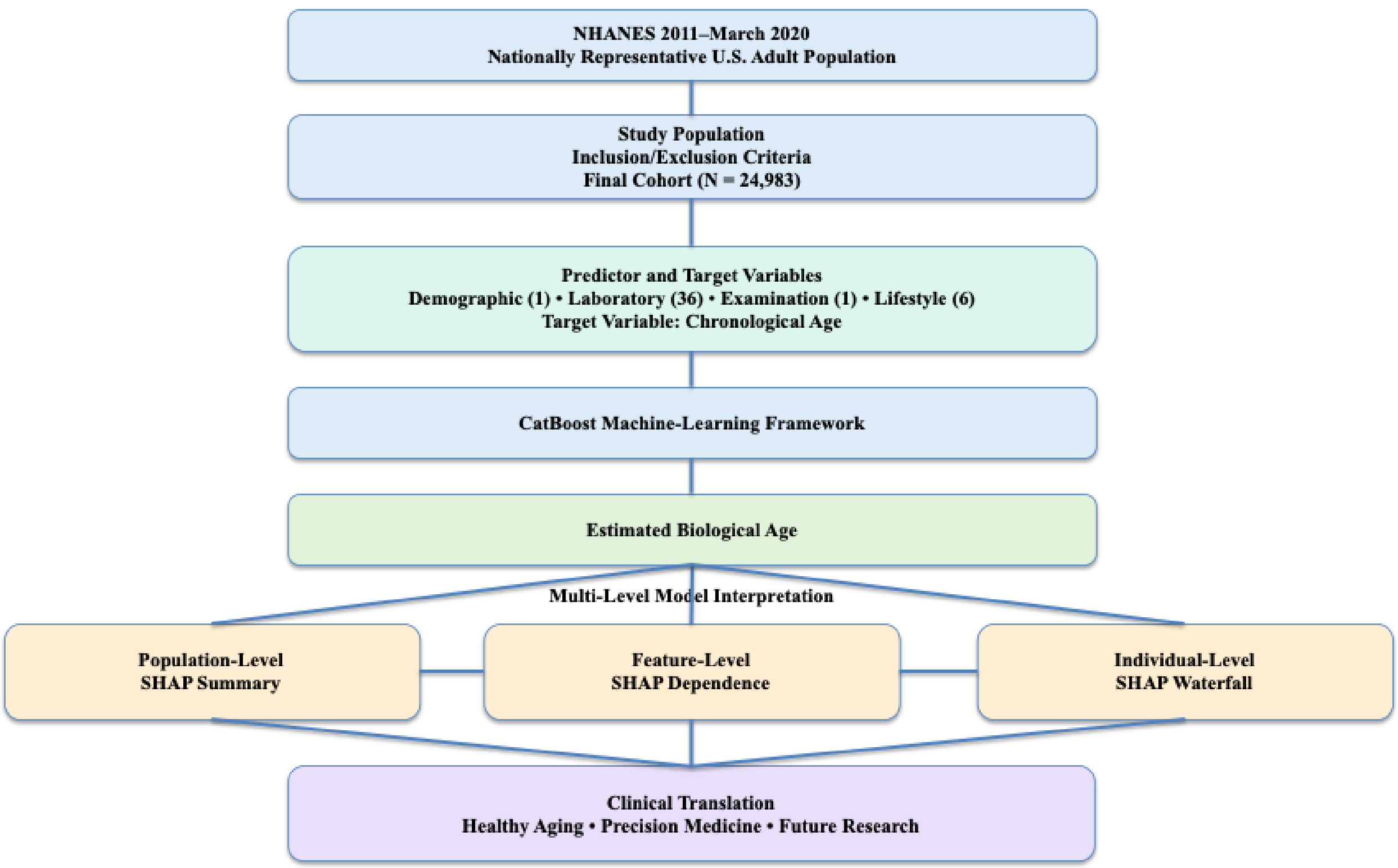

Figure 1. Overall Study Design and Analytical Framework for Clinically Interpretable Biological Age Estimation. Adults aged ≥18 years from NHANES 2011–March 2020 were included after application of predefined eligibility criteria. Forty-four demographic, laboratory, examination, and lifestyle predictors were incorporated into a CatBoost regression framework to estimate chronological age, with model-estimated age interpreted as biological age. Model interpretation was conducted at population, feature, and individual levels using SHAP.

### 2.3 Predictor and Target Variables

#### Target Variable

The response variable for supervised machine learning was CA, obtained from the NHANES demographic dataset. Following the framework of previously published first-generation BA models, CA was used as the supervised learning target, whereas the model-predicted age was interpreted as BA (Bae et al., 2021; Wood et al., 2019). To protect participant confidentiality and prevent deductive disclosure, NHANES top-coded ages: participant aged 80 or older is given a blanket age value of 80. The impact of the top-coding was investigated in the sensitivity analyses with and without age of 80.

**Predictor Variables**

A total of 44 pre-specified predictor variables representing four clinically relevant domains were included in model development based on literature review.

Demographic variable (n = 1): Gender.

Laboratory variables (n = 36): Thirty-six routinely measured blood biomarkers were selected based primarily on the interpretable BA model developed by Wood et al. (2019), while ensuring consistency across NHANES survey cycles. These biomarkers represented hematologic, metabolic, hepatic, renal, electrolyte, lipid, and inflammatory physiological systems.

Examination variable (n = 1): Body Mass Index (BMI)

Questionnaire variables (n = 6): Lifestyle and behavioral variables included workday sleep duration, daily sedentary time, history of trouble sleeping (i.e. 'Had trouble sleeping' in tables and figures), smoking history (i.e. 'Smoke Or Not' in tables and figures), vigorous recreational activity, and moderate recreational activity. These variables were selected because previous studies have demonstrated important associations between lifestyle behaviors and biological aging (Lu, 2024; Wang et al., 2022).

Detailed definitions of all predictor variables are provided in Table 1.

Table 1. Target Variable and Predictor Domains Included in the Clinically Interpretable Machine-Learning Framework.

| **Domain** | **Variable Type** | **Number of Variables** | **Examples** |
|---|---|---|---|
| Target | Chronological age | 1 | Age (year) |
| Demographic | Categorical | 1 | Gender |
| Laboratory | Continuous | 36 | BUN, Glucose, Albumin, ALT, AST |
| Examination | Continuous | 1 | Body mass index |
| Lifestyle/Questionnaire | Continuous | 2 | Workday sleep hours, Minutes sedentary activity |
| Lifestyle/Questionnaire | Categorical | 4 | Smoke Or Not, Vigorous recreational activity, Moderate recreational activity, Had trouble sleeping |

Footnote: A complete list of all 44 predictor variables is provided in Supplementary Table S1.

Responses of "Refused" and "Don't know" were treated as missing values. Yes/no questionnaire variables were coded as Yes = 1 and No = 0; gender was retained as a categorical predictor. Continuous variables retained their original measurement scales.

### 2.4 Model Development

A CatBoost regression model was developed to estimate CA using the 44 predictor variables. CatBoost was selected because it natively supports heterogeneous datasets containing both continuous and categorical variables through ordered target statistics while minimizing prediction bias and reducing overfitting (Prokhorenkova et al., 2018). Compared with conventional gradient boosting algorithms that require extensive categorical encoding, CatBoost provides an efficient framework for integrating heterogeneous clinical predictors with minimal preprocessing.

The analytical dataset was randomly divided into an 80% training dataset ($N$ = 19,986) and a 20% held-out testing dataset ($N$ = 4,997) using a fixed random seed (seed = 1000) to ensure reproducibility. Model development was performed exclusively using the training dataset, whereas model performance was evaluated using the held-out testing dataset.

Default CatBoost hyperparameters were used (loss function and evaluation metric: *RMSE;* random seed = 1000), with no additional hyperparameter tuning performed. Analyses were conducted in Python 3.10.2 using CatBoost 1.2.2. The trained CatBoost model generated predicted ages for all participants in the held-out testing dataset. Consistent with previous first-generation BA models, these predicted ages were interpreted as estimates of BA (Bae et al., 2021; Wood et al., 2019).

### 2.5 Model Interpretation

Model interpretation was performed using SHapley Additive exPlanations (SHAP, version 0.49.1), an explainable artificial intelligence approach that provides theoretically grounded feature attribution for machine-learning predictions (Lundberg & Lee, 2017). Quantitative evaluations of additive local explanation methods have further emphasized the importance of reproducible metrics for explainable artificial intelligence (Doumard et al., 2023). To facilitate clinically interpretable BA estimation, model interpretation was conducted at three complementary levels.

#### Population-Level Interpretation

Global feature importance was evaluated using SHAP summary plots generated from the training dataset. These plots simultaneously display feature importance, feature value distribution, and the direction of each predictor's influence on BA estimation.

#### Feature-Level Interpretation

SHAP dependence plots were generated for the most influential predictors to characterize nonlinear relationships between predictor values and their corresponding SHAP values. For each major predictor, three prespecified quantitative characteristics were calculated directly from the SHAP dependence data:

1. the observed predictor-value ranges among observations in the lowest 5% and highest 5% of SHAP values, representing the most negative and most positive contributions to the model-estimated BA, respectively;
2. the corresponding BA adjustment magnitudes, summarized as the SHAP-value ranges within the lowest 5% and highest 5% subsets; and
3. the SHAP zero-crossing point, conceptually defined as the predictor value at which the SHAP contribution crossed zero, operationally calculated as the mean predictor value among the 5% of observations with the smallest absolute SHAP contribution ($|SHAP| \leq$ 5th percentile of $|SHAP|$ for that predictor) rather than the single feature value nearest SHAP = 0, representing an approximate transition between net negative and net positive contribution to the BA estimate.

Building on the visual interpretation approach established by Wood et al. (2019), these prespecified characteristics were calculated algorithmically to provide reproducible quantitative interpretation of SHAP dependence plots. We applied the same SHAP zero-crossing point calculation to top 10 predictors, including both continuous and categorical variables; however, the resulting value for a categorical variable may not represent a meaningful inflection point. For categorical predictors, quantitative interpretation should be based on the observed discrete categories rather than a continuous zero-crossing calculation.

**Individual-Level Interpretation**

Individualized model interpretation was performed using SHAP waterfall plots. For selected participants from the held-out testing dataset, waterfall plots illustrated how each predictor contributed to the final BA estimate relative to the model's expected value. These individualized explanations complement the population-level SHAP analyses by demonstrating how specific clinical characteristics collectively influence BA estimation for an individual participant.

## 2.6 Statistical Analysis

Descriptive statistics were calculated for all predictor variables. Continuous variables are presented as means with standard deviations or medians with ranges where appropriate, whereas categorical variables are summarized as frequencies and percentages.

Model performance was evaluated by comparing predicted BA with CA in the held-out testing dataset. Pearson correlation coefficients and root mean squared error (*RMSE*) were calculated to assess predictive performance, consistent with previous BA studies (Bae et al., 2021; Wood et al., 2019).

All data preprocessing, model development, and statistical analyses were performed using Python. CatBoost was used for machine learning model development, and the SHAP Python library was used for explainable artificial intelligence analyses.

# 3. Results

## 3.1 Study Population

A total of 24,983 adults from NHANES 2011–March 2020 met the study eligibility criteria and comprised the analytical cohort (Figure 1). The cohort included 12,099 (48.4%) males and 12,884 (51.6%) females. Among the five categorical predictor variables, 40.9% of participants reported a history of smoking, 24.4% participated in vigorous recreational activities, and 41.2% participated in moderate recreational activities. Self-reported trouble sleeping (Had trouble sleeping) was reported by 26.5% of participants. Missingness for categorical predictors ranged from 0.0% to 1.2% (Table 2; Supplementary Table S2).

Table 2. Characteristics of the Analytical Cohort (NHANES 2011–March 2020).

| **Characteristic** | **Value** |
|---|---|
| Participants, n | 24,983 |
| Age, year, mean ± SD | 48.3 ± 18.4 |
| Age range, year | 18–80 |
| Male, n (%) | 12,099 (48.4) |
| Female, n (%) | 12,884 (51.6) |
| Smoking history, n (%) | 10,215 (40.9) |
| Vigorous recreational activity, n (%) | 6,089 (24.4) |
| Moderate recreational activity, n (%) | 10,305 (41.2) |
| Trouble sleeping, n (%) | 6,616 (26.5) |

The mean chronological age (CA) of the study population was 48.3 ± 18.4 years, with participants ranging from 18 to 80 years of age. The study included 39 continuous predictor variables representing hematologic, metabolic, hepatic, renal, lipid, electrolyte, anthropometric, and lifestyle characteristics. Missingness for individual continuous variables was generally low (≤3%) (Table 2; Supplementary Table S2).

The broad age distribution together with the diversity of demographic, laboratory, examination, and lifestyle characteristics provided a heterogeneous clinical dataset suitable for developing and evaluating a clinically interpretable biological age (BA) estimation framework.

### 3.2 Biological Age Estimation Performance

In the held-out testing dataset, model-estimated BA was strongly positively correlated with CA ($r = 0.798$, $P < 0.001$) (Figure 2). The fitted regression line is shown relative to the line of identity across the age spectrum; individual BA estimates varied around this relationship.

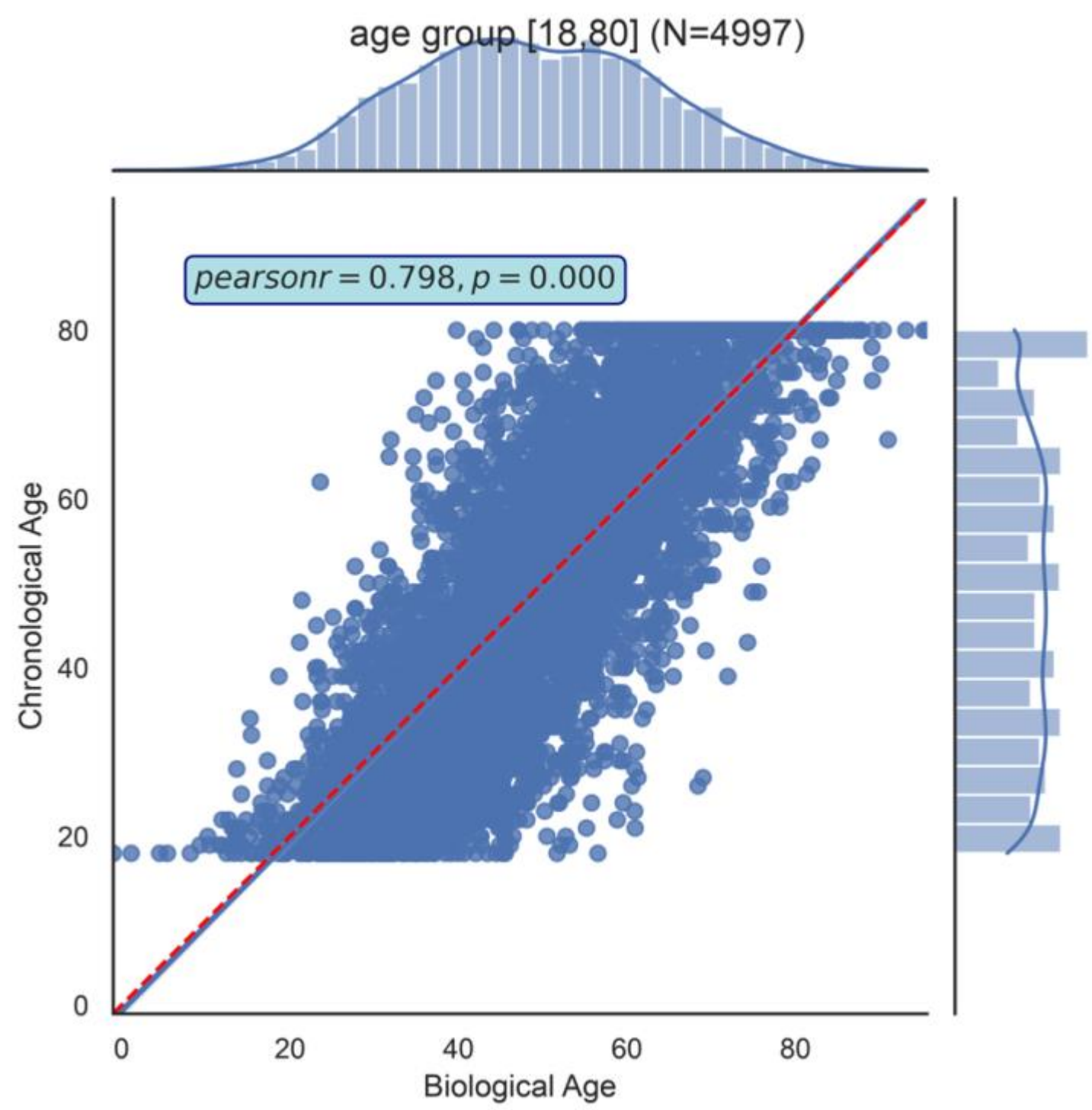

Figure 2. Relationship Between Chronological Age and Estimated Biological Age. The scatter plot displays chronological age and CatBoost-estimated biological age in the held-out testing dataset ($N = 4{,}997$). The fitted regression line (solid blue) is shown relative to the line of identity (dashed red). Marginal histograms display the distributions of chronological and estimated biological age. Model-estimated biological age was strongly correlated with chronological age (Pearson $r = 0.798$, $P < 0.001$).

The root mean squared error (*RMSE*) of the primary model was 11.0 years (Table 3). Variability in estimated BA was greater near the extremes of the chronological-age distribution and narrower among middle-aged participants (Figure 2).

Table 3. Performance of the CatBoost Framework for Biological Age Estimation in the Held-Out Testing Dataset.

| **Performance Metric** | **Value** |
|---|---|
| Training dataset (*N*) | 19,986 |
| Testing dataset (*N*) | 4,997 |
| Pearson correlation (*r*) | 0.798 |
| P value | <0.001 |
| Root mean squared error (*RMSE*), year | 11.0 |

Together, these performance results supported subsequent evaluation of the model at the population, feature, and individual interpretation levels.

### 3.3 Population-Level Model Interpretation

Population-level SHAP interpretation identified both laboratory biomarkers and categorical clinical variables among the highest-ranking contributors to BA estimation (Figure 3). Blood urea nitrogen (BUN) was the most influential predictor in the overall model, followed by serum glucose and vigorous recreational activity. Red cell distribution width (RDW), mean corpuscular volume (MCV), alanine aminotransferase (ALT), serum albumin, triglycerides, lymphocyte count, and aspartate aminotransferase (AST) also ranked among the most influential predictors (Figure 3).

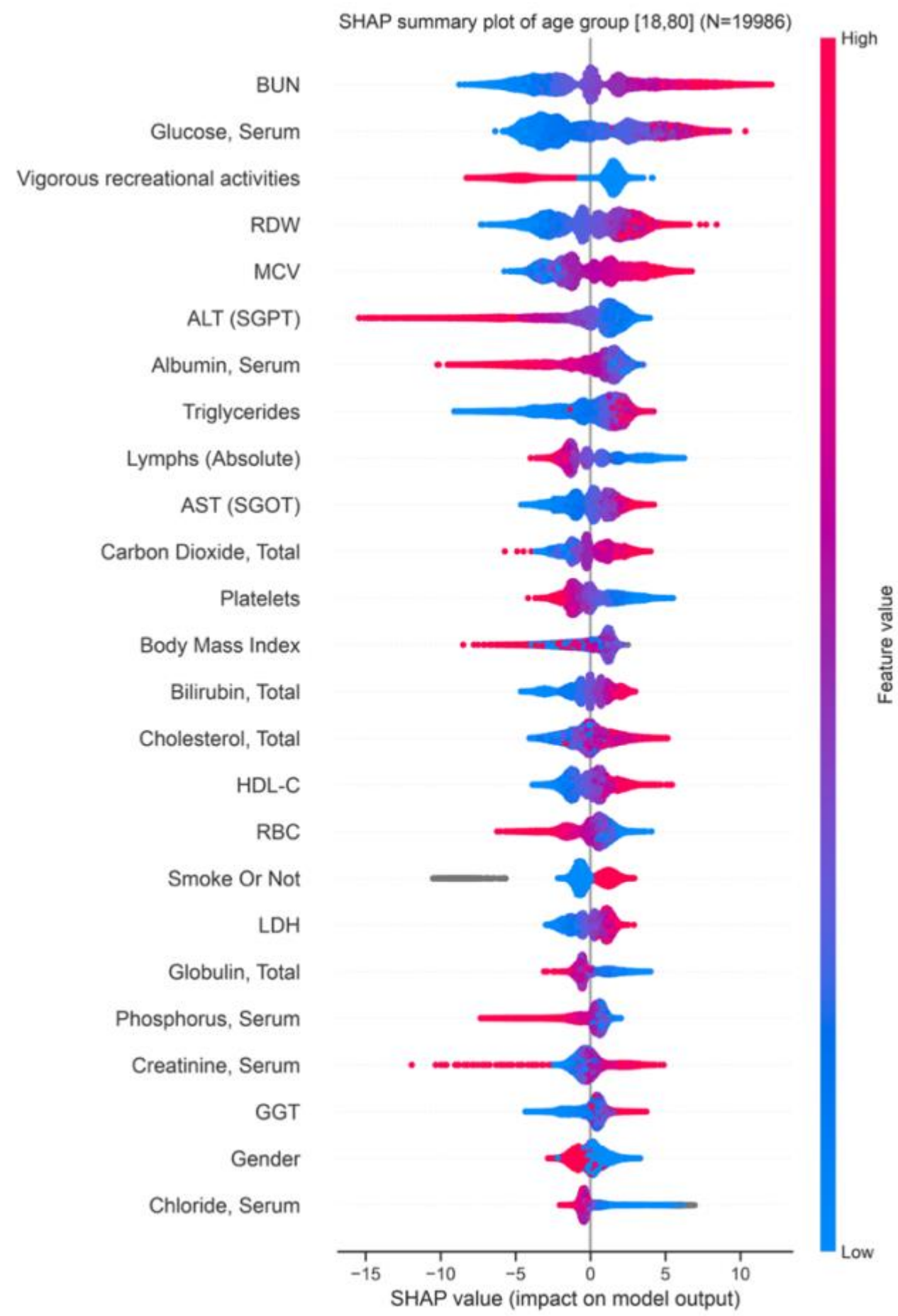

Figure 3. Population-Level Interpretation of Biological Age Estimation. The SHAP summary plot ranks the 25 most influential predictors in the training dataset (*N* = 19,986) and displays the direction and magnitude of their contributions to estimated biological age. Negative SHAP values decrease and positive SHAP values increase the model estimate. Point color represents feature value from low to high; gray points denote missing values. Laboratory biomarkers and categorical clinical predictors, including vigorous recreational activity, smoking history, and gender, contributed to population-level model interpretation.

The SHAP summary plot simultaneously illustrated feature importance and the direction of each predictor's contribution to BA estimation. Lower values of BUN and serum glucose were generally associated with negative SHAP values, indicating lower estimated BA, whereas higher values contributed positively to BA estimation (Figure 3). In contrast, participation in vigorous recreational activities was associated with negative SHAP values, reflecting lower estimated BA among physically active participants. Smoking history (Smoke Or Not) demonstrated the opposite pattern, contributing positively to BA estimation.

Notably, three categorical variables—vigorous recreational activity, smoking history, and gender—were ranked among the 25 most influential predictors alongside conventional laboratory biomarkers (Figure 3), demonstrating that heterogeneous clinical variables contributed to the model at the population level.

The SHAP summary plot also revealed substantial heterogeneity in predictor contributions across individuals, indicating that the influence of individual variables varied according to each participant's clinical profile. This observation motivated subsequent feature-level analyses to characterize the nonlinear relationships between major predictors and BA estimation.

### 3.4 Feature-Level Interpretation

Feature-level interpretation using SHAP dependence plots further characterized relationships between major predictors and BA estimation by quantifying predictor-specific value ranges, BA adjustment magnitudes, and zero-crossing points (Figure 4, Table 4). For each major predictor, SHAP values were examined across the observed feature range to identify predictor-value ranges associated with the lowest and highest BA contributions.

Building on previous visual interpretation of SHAP dependence plots, the present framework automatically quantified predictor-value ranges associated with the lowest 5% and highest 5% BA contributions, corresponding BA adjustment magnitudes, and predictor-specific SHAP zero-crossing points (Table 4).

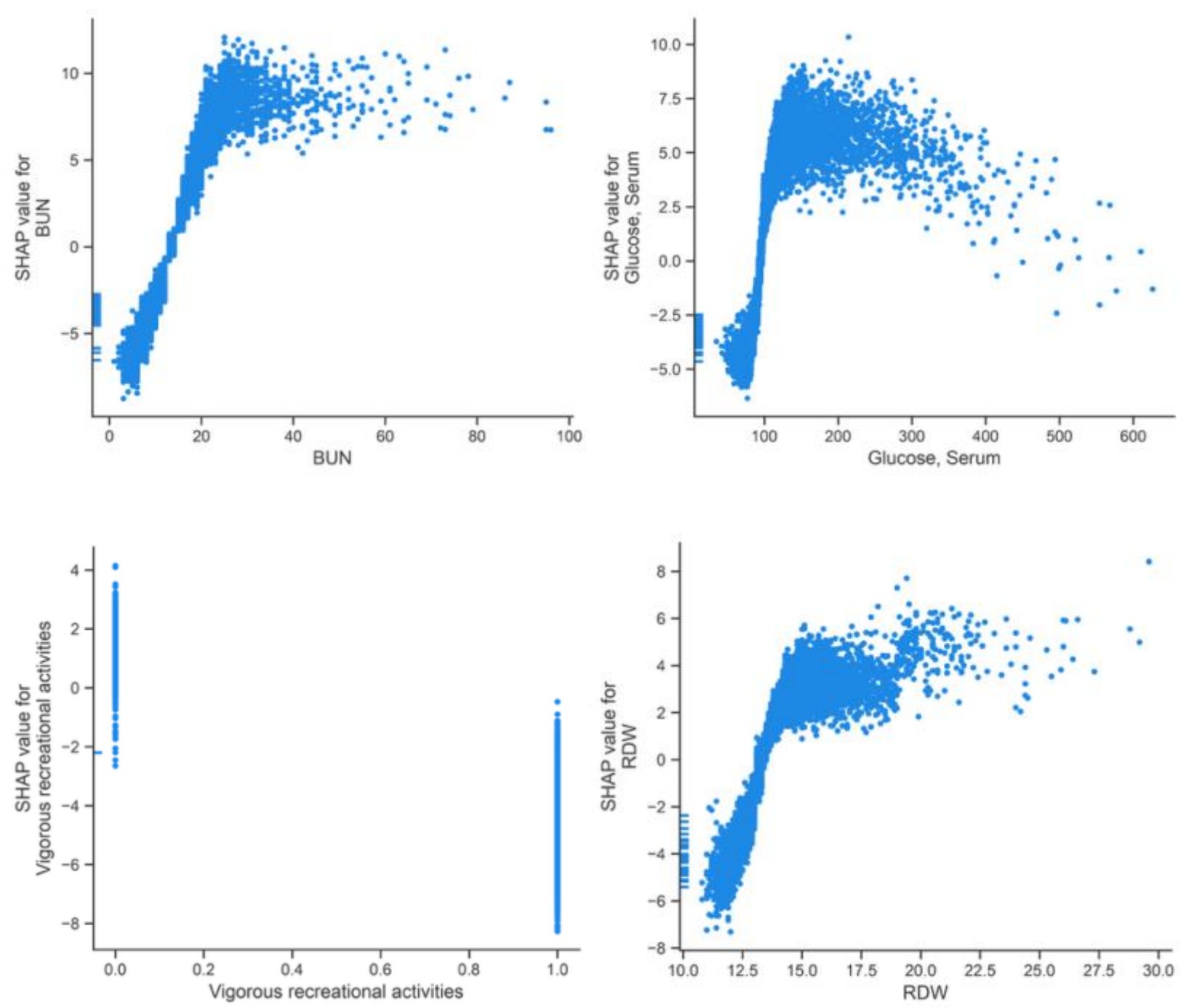

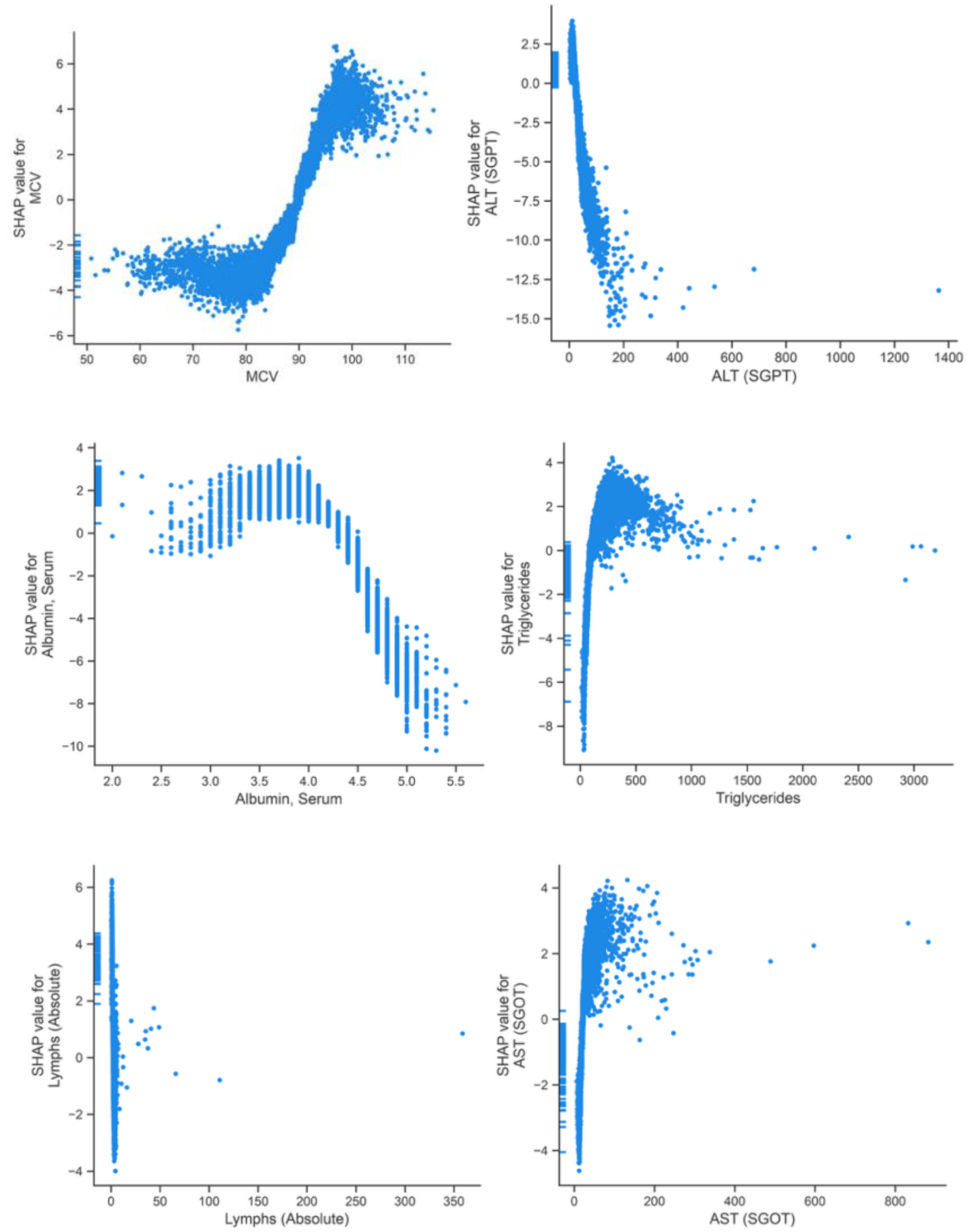

Figure 4. Automated Quantitative Characterization of SHAP Dependence Plots for Clinically Interpretable Biological Age Estimation. SHAP dependence plots illustrate predictor-specific relationships between feature values and contributions to model-estimated biological age. The automated procedure characterizes predictor-value ranges corresponding to the lowest and highest 5% of SHAP contributions, their associated biological age adjustment ranges, and SHAP zero-crossing points for continuous predictors.

Table 4. Automated Quantitative Interpretation of SHAP Dependence Plots for Clinically Interpretable Biological Age Estimation. Automated quantitative results are reported for the top 10 predictors in the CatBoost model based on the training dataset ($N = 19{,}986$). Predictor-value ranges and corresponding biological age adjustment magnitudes were derived from observations in the lowest and highest 5% of SHAP values for each predictor. The SHAP zero-crossing point represents the predictor value at which the SHAP contribution transitions between negative and positive contributions to the model-estimated biological age.

| Feature | Predictor-Value Range among Lowest 5% SHAP Values | BA Adjustment Magnitude among Lowest 5% SHAP Values (Years) | SHAP Zero-Crossing Point | Predictor-Value Range among Highest 5% SHAP Values | BA Adjustment Magnitude among Highest 5% SHAP Values (Years) |
|---|---|---|---|---|---|
| BUN | 1 - 9 mg/dL | -8.7 to -2.2 | 13.4 mg/dL | 20 – 95 mg/dL | 3.0 to 12.1 |
| Glucose, Serum | 48 - 83 mg/dL | -6.4 to -1.6 | 96.2 mg/dL | 112 - 356 mg/dL | 1.5 to 10.3 |
| Vigorous recreational activities | 1 | -8.3 to -0.5 | NA* | 0 | -2.6 to 4.2 |
| RDW | 10.8 - 12.7 % | -7.3 to -1.0 | 13.3% | 14.2 - 29.6 % | 0.9 to 8.4 |
| MCV | 57.6 - 84.8 fL | -5.7 to -1.2 | 89.7 fL | 94 - 113.8 fL | 1.8 to 6.8 |
| ALT (SGPT) | 43 - 1363 IU/L | -15.5 to -2.0 | 22.8 IU/L | 3 - 17 IU/L | 0.04 to 4.0 |
| Albumin, Serum | 4.7 - 5.6 g/dL | -10.2 to -2.2 | 4.3 g/dL | 2.1 - 4 g/dL | -1.1 to 3.5 |
| Triglycerides | 12 - 59 mg/dL | -9.1 to -1.4 | 110.4 mg/dL | 162 - 872 mg/dL | -1.7 to 4.2 |
| Lymphs (Absolute) | 2.6 - 5.6 1000 cells/uL | -4.0 to 3.2 | 2.0 1000 cells/uL | 0.2 - 1.4 1000 cells/uL | 1.3 to 6.3 |
| AST (SGOT) | 7 - 16 IU/L | -4.6 to -0.6 | 21.1 IU/L | 27 - 882 IU/L | -0.6 to 4.2 |

* Zero-crossing points are not applicable to binary categorical predictors.

Among all predictors, blood urea nitrogen (BUN) demonstrated the widest contribution to BA estimation. Lower BUN concentrations were associated with progressively more negative SHAP values, whereas higher BUN concentrations contributed positively to estimated BA (Figure 4). Quantitative analysis indicated that among participants with lowest 5% of SHAP values, BUN values ranged between 1 and 9 mg/dL and the corresponding BA adjustment magnitudes between -8.7 to -2.2 years; whereas among participants with highest 5% of SHAP values, BUN values ranged between 20 and 95 mg/dL and the corresponding BA adjustment magnitudes between 3.0 to 12.1 years. The SHAP zero-crossing point occurred at 13.4 mg/dL, where the SHAP value crossed zero (Table 4).

A similar nonlinear relationship was observed for serum glucose (Figure 4). Lower glucose concentrations were associated with more negative BA contributions, whereas increasing glucose values progressively contributed to higher BA estimates. The estimated SHAP zero-crossing point occurred at 96.2 mg/dL. Among participants with lowest 5% of SHAP values, the glucose values ranged between 48 and 83 mg/dL associated with the lowest BA contributions with adjustment magnitudes between -6.4 to -1.6 years (Table 4).

The dependence plot for vigorous recreational activity demonstrated a distinct pattern characteristic of categorical predictors (Figure 4). Participants reporting vigorous recreational activity generally exhibited negative SHAP values relative to those reporting no vigorous activity, indicating lower estimated BA. Unlike continuous laboratory biomarkers, the effect of this predictor was represented by two discrete categories corresponding to the presence or absence of vigorous recreational exercise.

Among additional high-ranking laboratory biomarkers, serum albumin demonstrated progressively lower BA estimates at higher concentrations, whereas alanine aminotransferase (ALT) exhibited a nonlinear relationship with BA estimation (Figure 4). Similar nonlinear patterns were observed for RDW, MCV, triglycerides, lymphocyte count, and AST (Table 4).

Collectively, these quantitative SHAP analyses extended interpretation beyond predictor ranking by identifying predictor-specific effect directions, value ranges associated with the lowest and highest BA contributions, adjustment magnitudes, and transition points.

### 3.5 Individual-Level Interpretation

To demonstrate individualized model interpretation, SHAP waterfall plots were generated for representative participants from the held-out testing dataset (Figure 5; Supplementary Table S3). These plots illustrate how individual predictor contributions collectively modify the model's expected BA to produce the final participant-specific BA estimate.

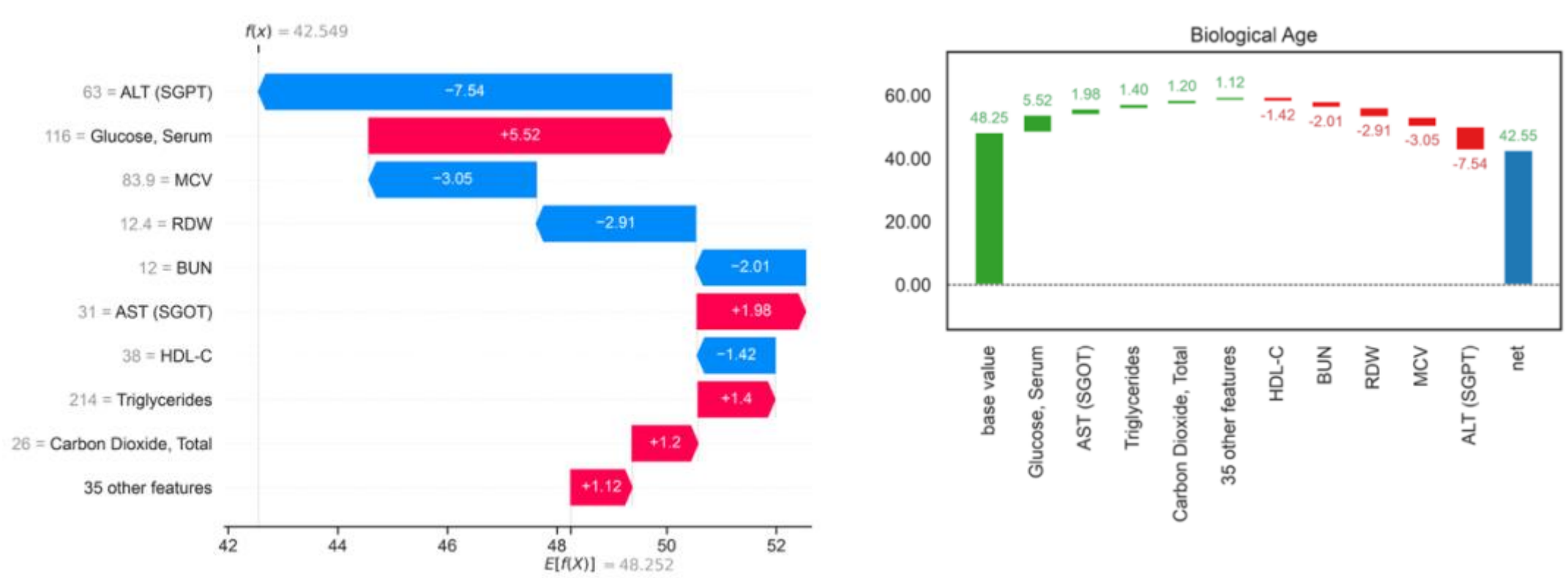

A B

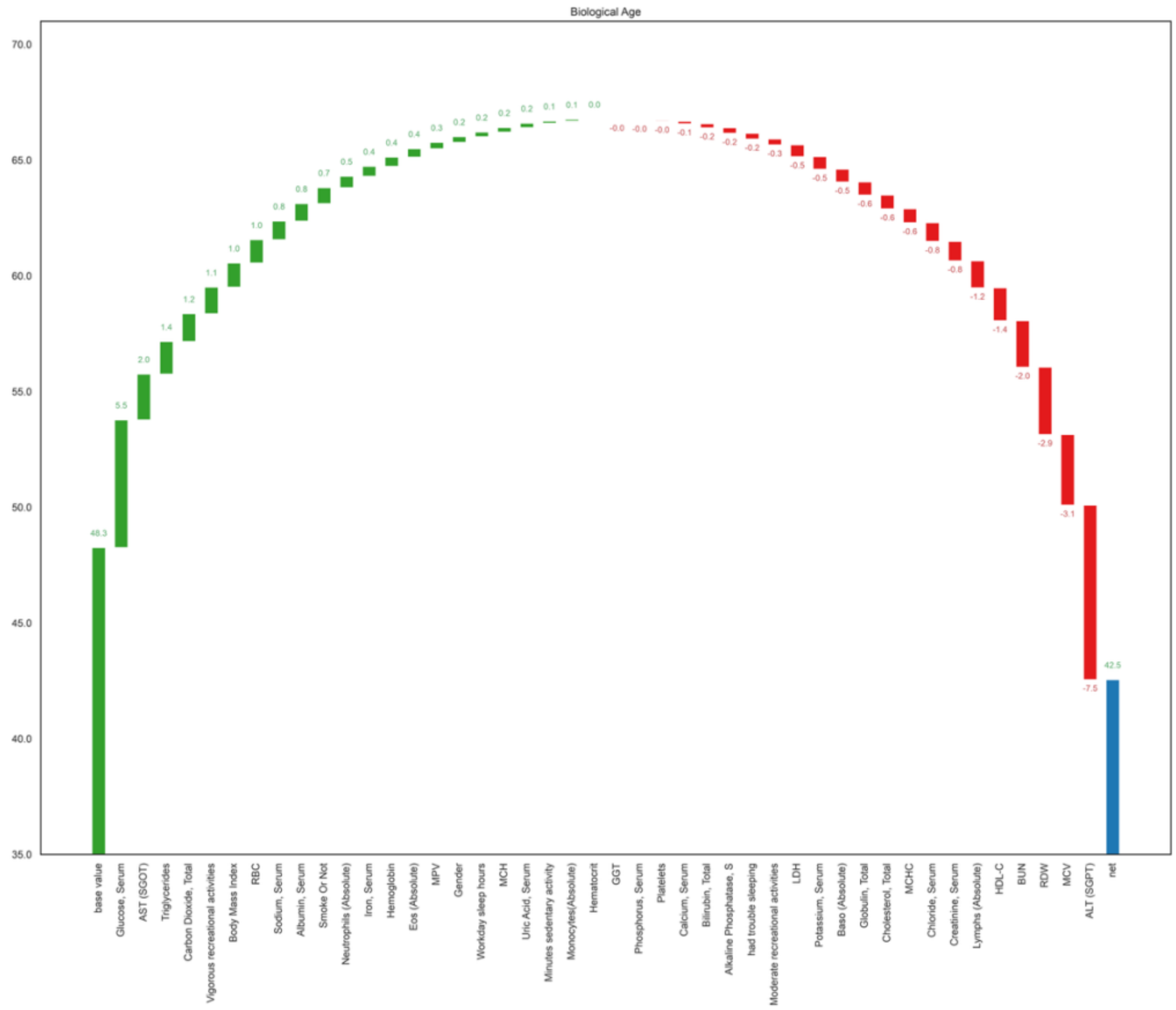

C

Figure 5. Individual-Level Interpretation of Biological Age Estimation. SHAP waterfall visualizations illustrate how participant-specific predictor contributions shift the model's expected value toward the final estimated biological age. Positive SHAP values increase and negative SHAP values decrease the estimate. Panels A–C present complementary representations of the same participant-level prediction. For the illustrated participant (D1; chronological age, 34 years), the model estimated a biological age of 42.5 years.

For the representative participant (Participant D1; CA = 34 years), the model estimated a BA of 42.5 years (Figure 5). The model's expected BA was 48.3 years, which was subsequently adjusted by the cumulative contributions of all predictor variables. The largest reductions in estimated BA were contributed by ALT, MCV, RDW, BUN, and HDL cholesterol, whereas serum glucose, AST, triglycerides, and carbon dioxide produced the largest increases in estimated BA (Figure 5).

Unlike the SHAP summary plot, which represents average feature importance across the study population, the waterfall plot demonstrated that the relative contribution of individual predictors varied substantially for a specific participant. For example, although BUN ranked as the most influential predictor at the population level, its contribution to Participant D1 was smaller than that of ALT because the participant's BUN concentration was located near the predictor's SHAP zero-crossing point (Figure 4; Figure 5; Supplementary Table S3).

These findings illustrate the complementary roles of population-level and individual-level model interpretation. Population-level analyses identify the predictors that generally contribute most to BA estimation across the study population, whereas individualized SHAP analyses quantify how each participant's unique clinical profile contributes to his or her estimated BA.

### 3.6 Secondary Analyses

Secondary analyses examined age-group-specific model behavior and the influence of missing participant-level predictor information on individualized interpretation.

Age-specific CatBoost models showed differences in predictor rankings and model performance between adults aged 18–64 and 65–80 years (Supplementary Figure S2; Supplementary Table S4). BUN remained among the highest-ranking predictors in both age-specific models. To examine the impacts of age top-coding in NHANES, sensitivity analyses were conducted for age groups 18–79 and 65–79. Minor impacts were observed (data on file and available upon request).

Exploratory individualized analyses also examined participants with missing predictor values. Although CatBoost natively handles missing values, waterfall explanations became less informative when many participant-level predictors were unavailable.

Overall, these secondary analyses suggest that predictor importance, model performance, and individualized explanations may vary by age group and by the completeness of participant-level clinical information.

## 4. Discussion

This study developed a clinically interpretable machine-learning framework for biological age (BA) estimation that integrates heterogeneous clinical data with explainable artificial intelligence to provide transparent interpretation at the population, feature, and individual levels (Figure 6). Building on previous interpretable BA models, the framework extends this line of research in three methodological aspects. First, it integrates heterogeneous continuous and categorical clinical variables within a single machine-learning framework. Second, it provides multi-level model interpretation, progressing from population-level SHAP summary plots to feature-level SHAP dependence analyses and individualized SHAP waterfall explanations. Third, it introduces an automated quantitative interpretation methodology for SHAP dependence plots, extending visual interpretation into reproducible quantitative analysis by identifying predictor-value ranges associated with the lowest 5% and highest 5% BA contributions (SHAP values), corresponding BA adjustment magnitudes, and SHAP zero-crossing points. Collectively, these innovations establish a clinically interpretable framework that bridges machine learning, explainable artificial intelligence, and biological aging research.

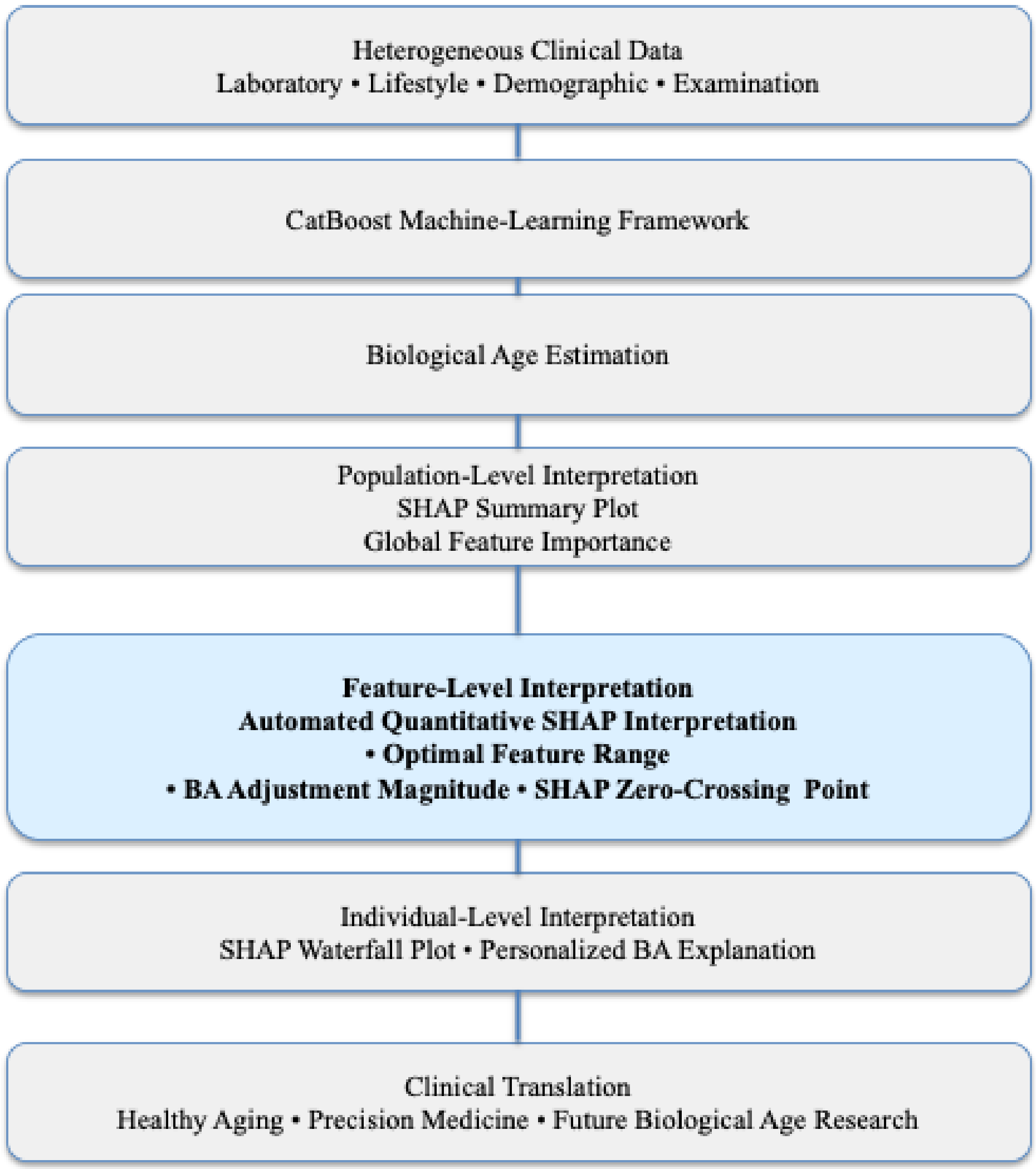

Figure 6. Conceptual Framework of Clinically Interpretable Machine-Learning for Biological Age Estimation Using Heterogeneous Clinical Data. The framework integrates heterogeneous demographic, laboratory, examination, and lifestyle predictors with CatBoost and organizes model interpretation across population, feature, and individual levels. Feature-level interpretation extends visual SHAP dependence analysis through prespecified quantitative characterization of extreme-contribution predictor-value ranges, biological age adjustment magnitudes, and SHAP zero-crossing points. The framework is intended to support reproducible interpretation of biological-age models rather than establish clinical reference ranges or treatment thresholds.

BA estimation has evolved considerably over the past decade. Early BA models primarily employed conventional statistical approaches using relatively small numbers of physiological biomarkers (Jee, 2016; Jia et al., 2017). The introduction of machine-learning algorithms subsequently enabled nonlinear modeling of complex biological relationships (Bae et al., 2021; Mamoshina et al., 2018). More recently, explainable artificial intelligence has been increasingly used to improve the transparency of machine-learning-based BA models (Bernard et al., 2023; Qiu et al., 2023). Wood et al. (2019) represented an important milestone by integrating XGBoost with SHAP to explain BA estimation using routinely measured laboratory biomarkers. Building upon this landmark foundation and subsequent work, the present framework extends interpretable BA estimation by integrating heterogeneous clinical variables, organizing interpretation across multiple levels, and adding automated quantitative SHAP dependence analysis.

One methodological extension of the present study is the integration of heterogeneous clinical data. Previous BA models have focused predominantly on continuous laboratory biomarkers, whereas demographic characteristics, lifestyle behaviors, and physical examination findings have often been underutilized or analyzed separately (Bae et al., 2021; Mamoshina et al., 2018; Wood et al., 2019). However, clinical data routinely contain heterogeneous information from multiple domains. By leveraging CatBoost's native handling of categorical predictors, the proposed framework simultaneously incorporates laboratory biomarkers, demographic characteristics, anthropometric measurements, and lifestyle variables within a single machine-learning model. The ranking of vigorous recreational activity, smoking history, and gender among the most influential predictors indicates that categorical demographic and lifestyle variables provided complementary predictive information alongside conventional laboratory biomarkers.

A second innovation is the introduction of multi-level model interpretation, which systematically connects population-level, feature-level, and individual-level understanding of BA estimation. Population-level SHAP summary plots identify the predictors that contribute most strongly to BA estimation across the study population. Feature-level SHAP dependence plots characterize nonlinear relationships between individual predictors and BA estimation. Individual-level SHAP waterfall plots subsequently explain how a participant's unique clinical profile contributes to the estimated BA. Rather than viewing these visualization methods independently, the present framework organizes them into complementary levels of interpretation that collectively enhance model transparency and facilitate clinically interpretable BA estimation.

A particularly important methodological contribution of this study is the extension of SHAP dependence plots from primarily visual interpretation toward reproducible quantitative characterization. Wood et al. (2019) demonstrated the value of SHAP dependence plots by visually estimating informative feature ranges, BA adjustment magnitude, and zero-crossing points, while Bernard et al. (2023) subsequently demonstrated quantitative SHAP-based metrics for personalized physiological age interpretation. Building upon these complementary foundations, the present framework applies an automated quantitative procedure directly to SHAP dependence plots to calculate three prespecified, model-derived characteristics for each major predictor: (1) the predictor-value ranges associated with the lowest and highest 5% BA contributions (SHAP values), (2) the corresponding BA adjustment magnitude associated with the lowest and highest 5% BA contributions (SHAP values), and (3) the predictor-specific SHAP zero-crossing point where the direction of contribution changes. This automated methodology is intended to improve reproducibility, reduce subjectivity associated with visual estimation, and facilitate objective comparison among predictors through a prespecified quantitative approach to SHAP dependence interpretation. Although developed within a BA framework, this methodology may also be applicable to other explainable machine-learning applications in biomedical research.

The proposed framework may also have implications for future clinical translation. Population-level interpretation identifies the variables most strongly associated with BA estimation across the study population. Quantitative feature-level interpretation provides model-derived predictor ranges and transition points that may facilitate biological interpretation of model behavior; these values should not be interpreted as clinical reference ranges or treatment thresholds. Individual-level interpretation further enables participant-specific explanation by quantifying the contribution of each predictor to an individual's estimated BA. Together, these complementary levels of interpretation move beyond prediction alone and provide a framework for future evaluation of clinically interpretable machine learning in personalized health assessment, healthy aging research, and precision medicine.

This study has several strengths. It utilized a large NHANES cohort with comprehensive laboratory, demographic, examination, and lifestyle information. The framework integrates heterogeneous clinical variables without extensive preprocessing of categorical predictors and combines prediction with transparent interpretation across multiple levels. Furthermore, the proposed automated quantitative SHAP interpretation methodology extends previous visual approaches by providing reproducible estimates of model-derived feature ranges, BA adjustment magnitudes, and SHAP zero-crossing points. Several limitations should also be acknowledged. Similar to previous first-generation BA models, chronological age (CA) served as the supervised learning target, with model-estimated age interpreted as BA; therefore, the model primarily captures age-related patterns in the included predictors and should not be interpreted as a direct measure of the underlying biological aging process. The predictors used in the model are prespecified based on limited literature review and investigators' judgement. This may preclude potential important predictors or include less important ones without a formal systematic literature review or a formal model selection procedure. The cross-sectional design also precludes assessment of within-person aging trajectories. Model performance was evaluated using a single held-out random split from the same analytical cohort rather than external validation, so predictive generalizability remains to be established. In addition, the present study focuses on BA estimation rather than prediction of aging-related clinical outcomes. Because the objective was model development and interpretation rather than estimation of nationally representative population parameters, the NHANES survey weights were not used in the analysis, and the findings should be interpreted as model-based results within the analytical cohort. External validation in independent cohorts and prospective evaluation against aging-related outcomes will be important before clinical application.

Future research should evaluate the generalizability of the proposed framework in independent populations, integrate additional clinical and molecular biomarkers, and extend the automated quantitative SHAP interpretation methodology to second-generation BA models and other explainable machine-learning applications. More broadly, the conceptual framework introduced in this study provides a foundation for evaluating clinically interpretable machine-learning approaches in biological aging research and, with appropriate external and prospective validation, their potential translation to clinical practice (Figure 6).

## 5. Conclusions

We developed a clinically interpretable machine-learning framework for biological age (BA) estimation that integrates heterogeneous clinical data with CatBoost and explainable artificial intelligence. Building on previous interpretable BA models, the framework combines heterogeneous continuous and categorical clinical variables, multi-level model interpretation, and automated quantitative SHAP dependence analysis. In particular, the quantitative approach extends visual interpretation by reproducibly characterizing predictor-value ranges associated with lower and higher BA contributions, BA adjustment magnitudes, and SHAP zero-crossing points. These methodological advances provide a foundation for clinically interpretable BA research and warrant further evaluation in independent populations and outcome-based aging models before clinical application.

## 6. Declarations

Acknowledgement and data availability: The authors thank the National Center for Health Statistics (NCHS) for providing the publicly accessible National Health and Nutrition Examination Survey (NHANES) data. The NHANES data are available at https://wwwn.cdc.gov/nchs/nhanes/.

Funding: Research reported in this publication was supported by the Graduate Education Program at Fei Tian College Northern Campus, Middletown, New York, USA. No external research funding was involved in supporting this work.

Conflict of interest: The authors declare that they have no conflict of interests.

Authors' contribution: ZC is the senior author contributed to supervision, conceptualization, methodology, writing – review & editing; YS conducted this MS degree thesis research and contributed to data curation, formal analysis, investigation, methodology, software, validation, visualization, writing – original draft; ZQ and KR contributed to conceptualization, resources, writing – review & editing. All authors gave their final approval and agreed to be accountable for all aspect of this work.

Supplementary material: For reproducibility and transparency, supplementary materials are provided with the submission.

Software availability: The Python codes following the framework developed in this work will be deposited/made available at an identified repository.

Generative-AI declarations: During the preparation of this manuscript, the authors used ChatGPT (OpenAI) to assist with review of manuscript organization, language refinement, clarity and readability, and consistency. The generative AI tool was not used to generate or analyze the study data or to make scientific conclusions. All AI-assisted content was critically reviewed, verified, and edited by the authors. The authors take full responsibility for the scientific content, accuracy, integrity, and final version of the manuscript.

## 8. Supplement Materials

Supplementary Table S1. Complete List of the Target Variable and 44 Predictor Variables Included in the CatBoost Machine-Learning Framework.

| **Domain** | **Variable** | **NHANES Variable Code** | **Type** | **Unit/Response Categories** |
|---|---|---|---|---|
| Target | Chronological age | RIDAGEYR | Continuous | years |
| Demographic | Gender | RIAGENDR | Categorical | Male/Female |
| Laboratory | Lymphs (Absolute) | LBDLYMNO | Continuous | 1000 cells/uL |
| Laboratory | Monocytes (Absolute) | LBDMONO | Continuous | 1000 cells/uL |
| Laboratory | Neutrophils (Absolute) | LBDNENO | Continuous | 1000 cell/uL |
| Laboratory | Eos (Absolute) | LBDEONO | Continuous | 1000 cells/uL |
| Laboratory | Baso (Absolute) | LBDBANO | Continuous | 1000 cells/uL |
| Laboratory | RBC | LBXRBCSI | Continuous | million cells/uL |
| Laboratory | Hemoglobin | LBXHGB | Continuous | g/dL |
| Laboratory | Hematocrit | LBXHCT | Continuous | % |
| Laboratory | MCV | LBXMCVSI | Continuous | fL |
| Laboratory | MCHC | LBXMC | Continuous | g/dL |
| Laboratory | MCH | LBXMCHSI | Continuous | pg |
| Laboratory | RDW | LBXRDW | Continuous | % |
| Laboratory | Platelets | LBXPLTSI | Continuous | 1000 cells/uL |
| Laboratory | MPV | LBXMPSI | Continuous | fL |
| Laboratory | HDL-C | LBDHDD | Continuous | mg/dL |
| Laboratory | ALT (SGPT) | LBXSATSI | Continuous | IU/L |
| Laboratory | Albumin, Serum | LBXSAL | Continuous | g/dL |
| Laboratory | Alkaline Phosphatase S | LBXSAPSI | Continuous | IU/L |
| Laboratory | AST (SGOT) | LBXSASSI | Continuous | IU/L |
| Laboratory | Carbon Dioxide, Total | LBXSC3SI | Continuous | mmol/L |
| Laboratory | BUN | LBXSBU | Continuous | mg/dL |
| Laboratory | Chloride, Serum | LBXSCLSI | Continuous | mmol/L |
| Laboratory | Creatinine, Serum | LBXSCR | Continuous | mg/dL |
| Laboratory | Globulin, Total | LBXSGB | Continuous | g/dL |
| Laboratory | Glucose, Serum | LBXSGL | Continuous | mg/dL |
| Laboratory | GGT | LBXSGTSI | Continuous | IU/L |
| Laboratory | Iron, Serum | LBXSIR | Continuous | ug/dL |
| Laboratory | LDH | LBXSLDSI | Continuous | IU/L |
| Laboratory | Phosphorus, Serum | LBXSPH | Continuous | mg/dL |
| Laboratory | Potassium, Serum | LBXSKSI | Continuous | mmol/L |
| Laboratory | Sodium, Serum | LBXSNASI | Continuous | mmol/L |
| Laboratory | Bilirubin, Total | LBXSTB | Continuous | mg/dL |
| Laboratory | Calcium, Serum | LBXSCA | Continuous | mg/dL |
| Laboratory | Triglycerides | LBXSTR | Continuous | mg/dL |
| Laboratory | Uric Acid, Serum | LBXSUA | Continuous | mg/dL |
| Laboratory | Cholesterol, Total | LBXTC | Continuous | mg/dL |
| Examination | Body Mass Index | BMXBMI | Continuous | kg/m² |
| Questionnaire | Workday sleep hours | SLD010H/SLD012* | Continuous | hour/day |
| Questionnaire | Minutes sedentary activity | PAD680 | Continuous | minute/day |
| Questionnaire | Smoking history | SMQ020 | Categorical | Yes/No |
| Questionnaire | Trouble sleeping | SLQ050 | Categorical | Yes/No |
| Questionnaire | Vigorous recreational activity | PAQ650 | Categorical | Yes/No |
| Questionnaire | Moderate recreational activity | PAQ665 | Categorical | Yes/No |

*SLD010H: 'Workday sleep hours' in NHANES 2011-2012 and NHANES 2013-2014. SLD012: 'Workday sleep hours' in NHANES 2015-2016 and NHANES 2017-2020.

Supplementary Table S2. Summary of 39 Continuous and 5 Categorical Variables (N=24,983).

| **Continuous Variable** | **Mean** | **SD** | **Minimum** | **Median** | **Maximum** | **Missing, %** |
|---|---|---|---|---|---|---|
| Chronological Age (years) | 48.3 | 18.4 | 18 | 48 | 80 | 0 |
| Lymphs (Absolute) (1000 cells/uL) | 2.2 | 2.6 | 0.2 | 2.1 | 358.8 | 0.3 |
| Monocytes(Absolute) (1000 cells/uL) | 0.6 | 0.2 | 0 | 0.5 | 10.2 | 0.3 |
| Neutrophils (Absolute) (1000 cell/uL) | 4.2 | 1.7 | 0.1 | 4 | 35.2 | 0.3 |
| Eos (Absolute) (1000 cells/uL) | 0.2 | 0.2 | 0 | 0.2 | 4.3 | 0.3 |
| Baso (Absolute) (1000 cells/uL) | 0 | 0.1 | 0 | 0 | 0.8 | 0.3 |
| RBC (million cells/uL) | 4.7 | 0.5 | 1.7 | 4.7 | 8.3 | 0.2 |
| Hemoglobin (g/dL) | 13.9 | 1.6 | 5.4 | 14 | 19.9 | 0.2 |
| Hematocrit (%) | 41.4 | 4.3 | 17.9 | 41.5 | 59.1 | 0.2 |
| MCV (fL) | 88.8 | 6.1 | 35.4 | 89.3 | 115.6 | 0.2 |
| MCHC (g/dL) | 33.6 | 1.1 | 25.1 | 33.7 | 69.6 | 0.1 |
| MCH (pg) | 29.9 | 2.5 | 12.4 | 30.2 | 74.5 | 0.2 |
| RDW (%) | 13.7 | 1.4 | 10.8 | 13.4 | 29.8 | 0.2 |
| Platelets (1000 cells/uL) | 240.3 | 62.7 | 8 | 234 | 818 | 0.2 |
| MPV (fL) | 8.4 | 0.9 | 5.4 | 8.3 | 15.1 | 0.2 |
| HDL-C (mg/dL) | 53.2 | 16 | 5 | 51 | 226 | 1.9 |
| ALT (SGPT) (IU/L) | 23.9 | 20.4 | 2 | 19 | 1363 | 2.3 |
| Albumin, Serum (g/dL) | 4.2 | 0.4 | 2 | 4.2 | 5.6 | 2.2 |
| Alkaline Phosphatase S (IU/L) | 71.7 | 26 | 7 | 68 | 907 | 2.3 |
| AST (SGOT) (IU/L) | 24.3 | 16.7 | 6 | 22 | 882 | 2.4 |
| Carbon Dioxide, Total (mmol/L) | 25.1 | 2.3 | 14 | 25 | 43 | 2.2 |
| BUN (mg/dL) | 14 | 6.1 | 1 | 13 | 96 | 2.3 |
| Chloride, Serum (mmol/L) | 103 | 3.1 | 70 | 103 | 119 | 2.2 |
| Creatinine, Serum (mg/dL) | 0.9 | 0.5 | 0.2 | 0.8 | 17.4 | 2.2 |
| Globulin, Total (g/dL) | 2.9 | 0.5 | 0.7 | 2.9 | 7 | 2.3 |
| Glucose, Serum (mg/dL) | 103.1 | 39.7 | 19 | 93 | 626 | 2.2 |
| GGT (IU/L) | 29.2 | 46 | 2 | 19 | 2394 | 2.3 |
| Iron, Serum (ug/dL) | 84.1 | 35.8 | 2 | 80 | 557 | 2.4 |
| LDH (IU/L) | 138 | 34.9 | 32 | 133 | 1274 | 3 |
| Phosphorus, Serum (mg/dL) | 3.7 | 0.6 | 1 | 3.7 | 10.9 | 2.2 |
| Potassium, Serum (mmol/L) | 4 | 0.4 | 2.3 | 4 | 7.3 | 2.3 |
| Sodium, Serum (mmol/L) | 139.6 | 2.5 | 102 | 140 | 161 | 2.2 |
| Bilirubin, Total (mg/dL) | 0.6 | 0.3 | 0 | 0.5 | 7.1 | 2.3 |
| Calcium, Serum (mg/dL) | 9.4 | 0.4 | 6.4 | 9.3 | 14.8 | 2.4 |
| Triglycerides (mg/dL) | 146.6 | 124.5 | 12 | 115 | 6057 | 2.3 |
| Uric Acid, Serum (mg/dL) | 5.4 | 1.4 | 0.4 | 5.3 | 18 | 2.3 |
| Cholesterol, Total (mg/dL) | 187.7 | 41.6 | 59 | 184 | 813 | 1.9 |
| Workday sleep hours | 7.3 | 1.6 | 2 | 7 | 14.5 | 0.5 |
| Body Mass Index (kg/m**2) | 29.3 | 7.3 | 13.4 | 28.1 | 92.3 | 1.4 |
| Minutes sedentary activity | 368.3 | 203.3 | 0 | 360 | 1380 | 0.7 |

| **Categorical Variable** | **Male or Yes, n (%)** | **Female or No, n (%)** | **Missing, n (%)** |
|---|---|---|---|
| Gender | 12,099 (48%) | 12,884 (52%) | 0 (0%) |
| Had trouble sleeping | 6,616 (26%) | 18,356 (73%) | 11 (0.04%) |
| Smoke Or Not | 10,215 (41%) | 14,468 (58%) | 300 (1.2%) |
| Vigorous recreational activities | 6,089 (24%) | 18,893 (76%) | 1 (0.004%) |
| Moderate recreational activities | 10,305 (41%) | 14,672 (59%) | 6 (0.02%) |

Supplementary Table S3. Clinical Profiles of Representative Participants Used for Individual-Level SHAP Interpretation.

| **Feature** | **Representative Participant** | | | |
|---|---|---|---|---|
| | **D1** | **D2** | **D3** | **D4** |
| Chronological Age (years) | 34.00 | 64.00 | 72.00 | 54.00 |
| Lymphs (Absolute) (1000 cells/uL) | 2.20 | 3.10 | 2.50 | 1.50 |
| Monocytes(Absolute) (1000 cells/uL) | 0.60 | 0.40 | 0.90 | 0.50 |
| Neutrophils (Absolute) (1000 cell/uL) | 4.10 | 4.20 | 3.20 | 3.30 |
| Eos (Absolute) (1000 cells/uL) | 0.20 | 0.20 | 0.10 | 0.50 |
| Baso (Absolute) (1000 cells/uL) | 0.00 | 0.20 | 0.00 | 0.10 |
| RBC (million cells/uL) | 4.34 | 4.75 | 4.40 | 5.16 |
| Hemoglobin (g/dL) | 12.60 | 11.80 | 13.80 | 15.50 |
| Hematocrit (%) | 36.40 | 36.10 | 41.80 | 45.70 |
| MCV (fL) | 83.90 | 76.10 | 94.90 | 88.60 |
| MCHC (g/dL) | 34.50 | 32.50 | 33.10 | 33.80 |
| MCH (pg) | 29.00 | 24.80 | 31.40 | 30.00 |
| RDW (%) | 12.40 | 15.10 | 13.90 | 13.80 |
| Platelets (1000 cells/uL) | 230.00 | 316.00 | 262.00 | 156.00 |
| MPV (fL) | 8.10 | 8.00 | 8.60 | 9.00 |
| HDL-C (mg/dL) | 38.00 | 53.00 | 85.00 | NA |
| ALT (SGPT) (IU/L) | 63.00 | 23.00 | 18.00 | NA |
| Albumin, Serum (g/dL) | 4.20 | 4.10 | 3.80 | NA |
| Alkaline Phosphatase, S (IU/L) | 39.00 | 99.00 | 80.00 | NA |
| AST (SGOT) (IU/L) | 31.00 | 21.00 | 21.00 | NA |
| Carbon Dioxide, Total (mmol/L) | 26.00 | 22.00 | 22.00 | NA |
| BUN (mg/dL) | 12.00 | 13.00 | 13.00 | NA |
| Chloride, Serum (mmol/L) | 106.00 | 104.00 | 103.00 | NA |
| Creatinine, Serum (mg/dL) | 0.65 | 0.92 | 0.96 | NA |
| Globulin, Total (g/dL) | 2.90 | 3.40 | 2.80 | NA |
| Glucose, Serum (mg/dL) | 116.00 | 82.00 | 83.00 | NA |
| GGT (IU/L) | 17.00 | 18.00 | 164.00 | NA |
| Iron, Serum (ug/dL) | 72.00 | 61.00 | 124.00 | NA |
| LDH (IU/L) | 122.00 | 176.00 | 131.00 | NA |
| Phosphorus, Serum (mg/dL) | 2.60 | 4.20 | 3.40 | NA |
| Potassium, Serum (mmol/L) | 3.50 | 4.10 | 4.60 | NA |
| Sodium, Serum (mmol/L) | 142.00 | 144.00 | 139.00 | NA |
| Bilirubin, Total (mg/dL) | 0.50 | 0.10 | 0.30 | NA |
| Calcium, Serum (mg/dL) | 9.40 | 8.60 | 9.00 | NA |
| Triglycerides (mg/dL) | 214.00 | 153.00 | 116.00 | NA |
| Uric Acid, Serum (mg/dL) | 6.20 | 6.90 | 7.80 | NA |
| Cholesterol, Total (mg/dL) | 168.00 | 222.00 | 206.00 | NA |
| Body Mass Index (kg/m**2) | 31.40 | 28.40 | 18.90 | 28.90 |
| Minutes sedentary activity | 240.00 | 180.00 | 60.00 | 180.00 |
| Workday sleep hours | 7.00 | 7.00 | 5.00 | 5.50 |
| Had trouble sleeping | 0.00 | 0.00 | 0.00 | 0.00 |
| Smoke Or Not | 1.00 | 0.00 | 0.00 | 0.00 |
| Vigorous recreational activities | 0.00 | 0.00 | 0.00 | 0.00 |
| Moderate recreational activities | 1.00 | 0.00 | 0.00 | 0.00 |
| Gender | 0.00 | 0.00 | 1.00 | 1.00 |

Supplementary Table S4. Performance of the Overall and Age-Group-Specific CatBoost Models. Age-group analyses were exploratory. Performance metrics across restricted age ranges should not be interpreted as direct evidence of superiority of one model over another.

| **Age group CatBoost model** | **[18,80]** | **[18, 65)** | **[65,80]** |
|---|---|---|---|
| Participant number (*N*) | 24983 | 19336 | 5647 |
| Pearson correlation coefficient (*r*) | 0.798 | 0.719 | 0.537 |
| Root mean squared error (*RMSE*), year | 11.0 | 9.7 | 4.6 |
| *P-value* | <0.001 | <0.001 | <0.001 |
| Predicted BA for D1 (CA=34) (year) | 42.5 | 37.5 | 69.2 |
| Predicted BA for D2 (CA=64) (year) | 48.6 | 49.2 | 68.9 |
| Predicted BA for D3 (CA=72) (year) | 53.6 | 52.0 | 73.0 |
| Predicted BA for D4 (CA=54) (year) | 55.0 | 49.9 | 72.8 |

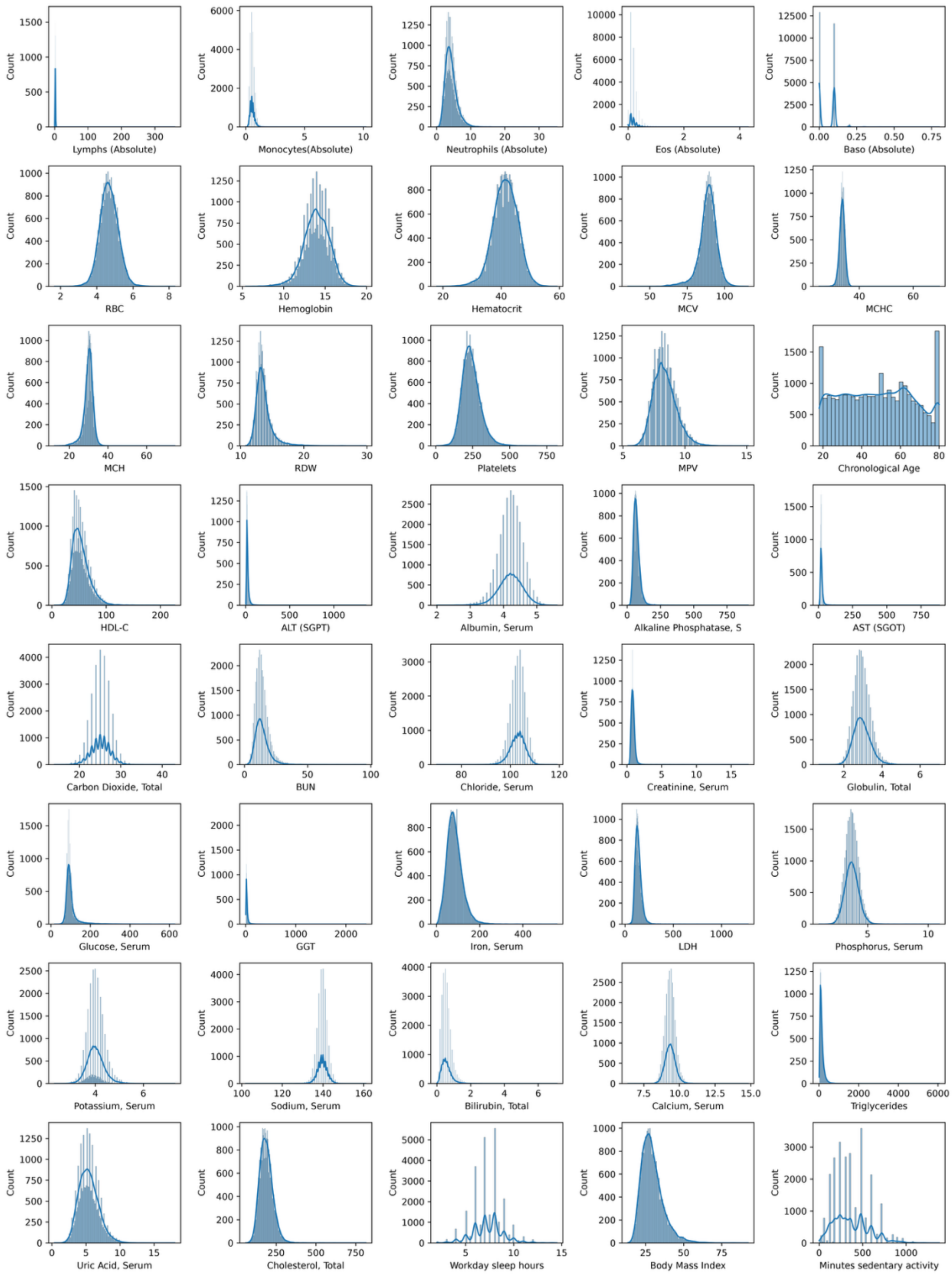
Count
Lymphs (Absolute)
Monocytes(Absolute)
Neutrophils (Absolute)
Eos (Absolute)
Baso (Absolute)
RBC
Hemoglobin
Hematocrit
MCV
MCHC
MCH
RDW
Platelets
MPV
Chronological Age
HDL-C
ALT (SGPT)
Albumin, Serum
Alkaline Phosphatase, S
AST (SGOT)
Carbon Dioxide, Total
BUN
Chloride, Serum
Creatinine, Serum
Globulin, Total
Glucose, Serum
GGT
Iron, Serum
LDH
Phosphorus, Serum
Potassium, Serum
Sodium, Serum
Bilirubin, Total
Calcium, Serum
Triglycerides
Uric Acid, Serum
Cholesterol, Total
Workday sleep hours
Body Mass Index
Minutes sedentary activity

Supplementary Figure S1. Distributions of Continuous Predictors and Chronological Age in the Analytical Cohort (*N* = 24,983). Histograms display the observed distributions of continuous predictor variables and the chronological-age target variable included in model development.

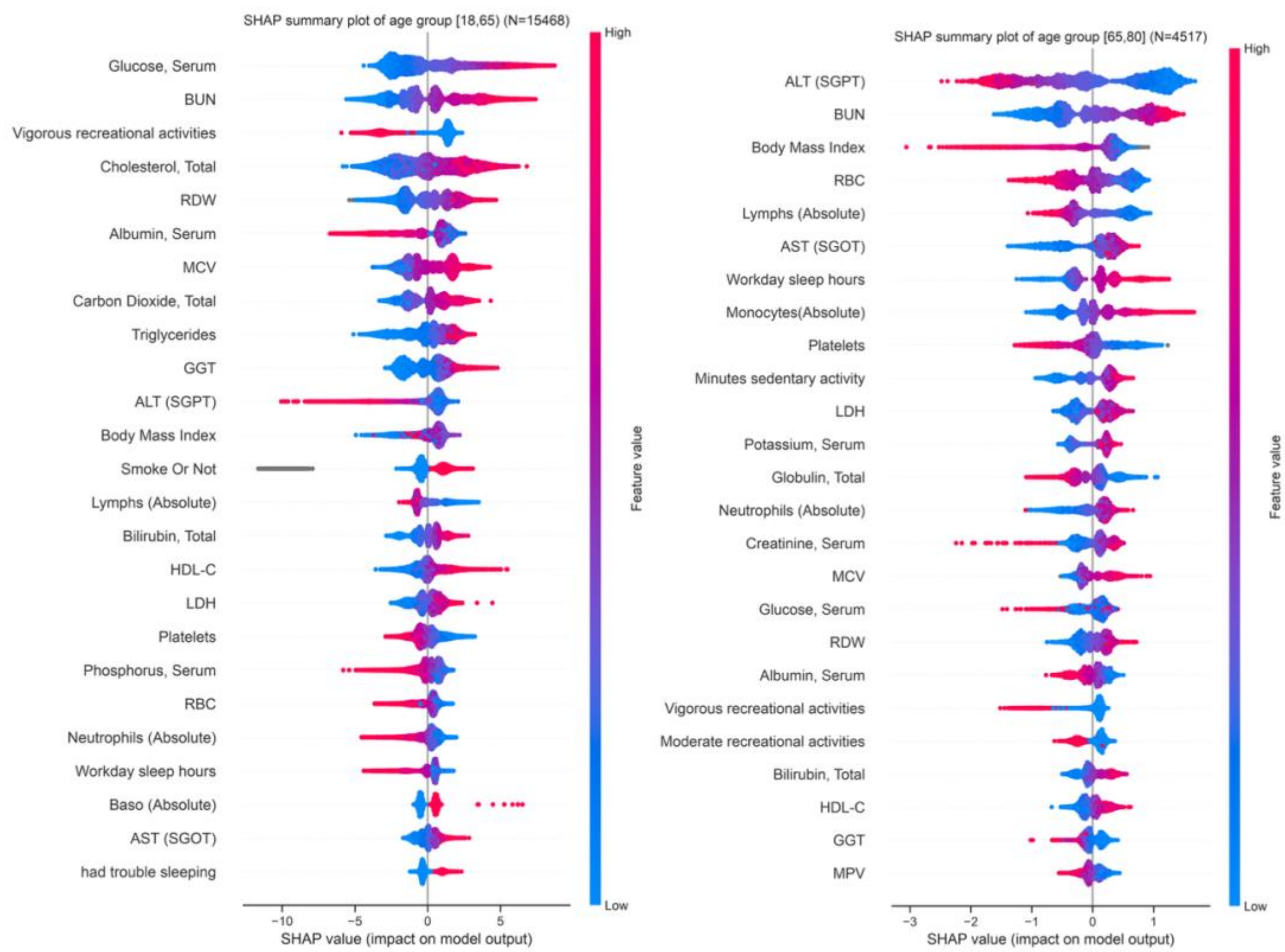

A B

Supplementary Figure S2. Population-Level SHAP Interpretation in Age-Group-Specific CatBoost Models. (A) Adults aged 18–64 years. (B) Adults aged 65–80 years. SHAP summary plots display predictor importance and direction of contribution within each age-group-specific model. Negative SHAP values decrease and positive SHAP values increase model-estimated

biological age; point color represents feature value from low to high, and gray points denote missing values. These analyses were exploratory.

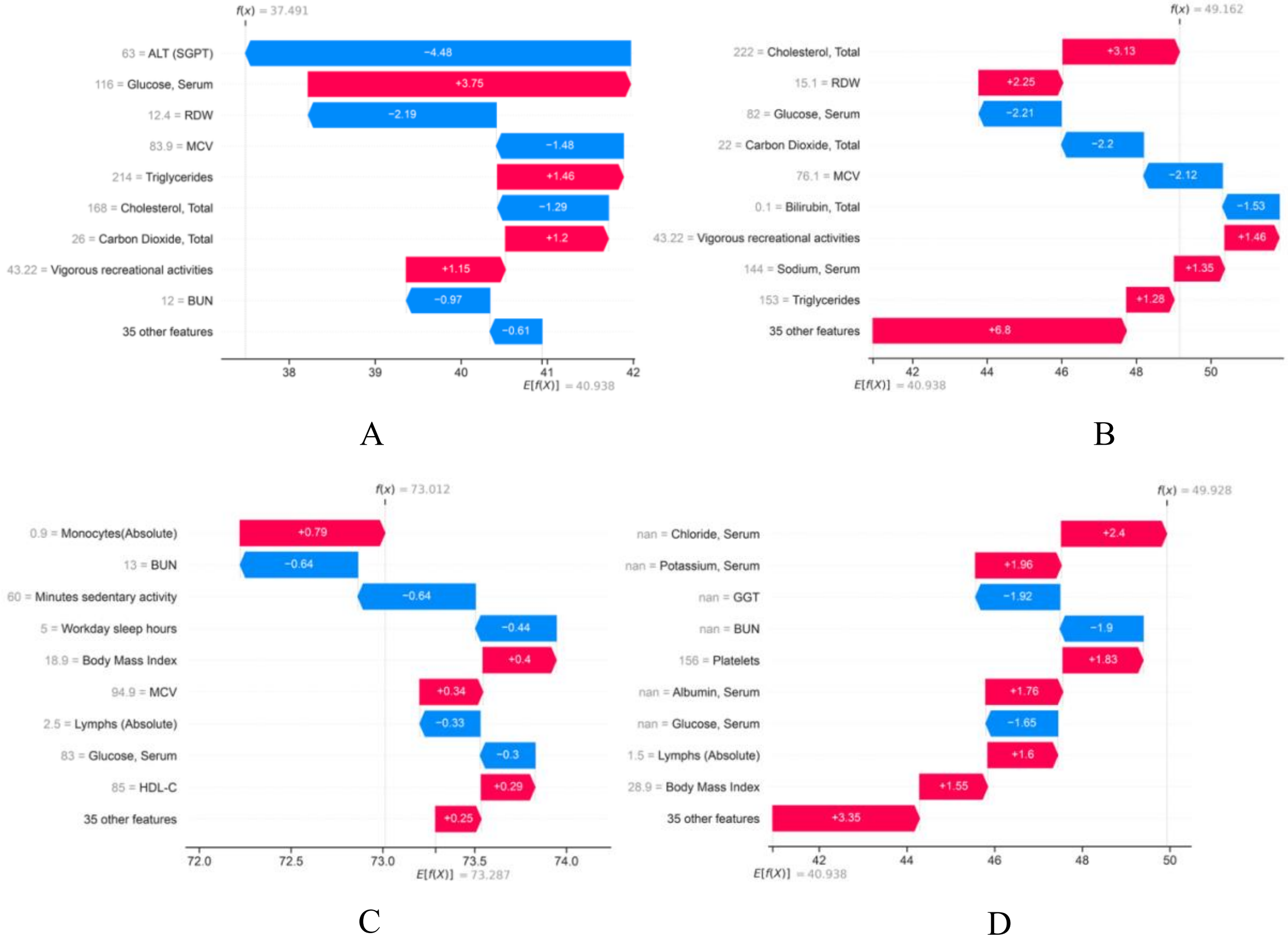

Supplementary Figure S3. Individual-Level SHAP Interpretations from Age-Group-Specific CatBoost Models. Waterfall plots illustrate participant-specific predictor contributions for representative participants evaluated using the age-group-specific model corresponding to their chronological age: (A) D1, CA = 34 years, 18–64-year model; (B) D2, CA = 64 years, 18–64-year model; (C) D3, CA = 72 years, 65–80-year model; and (D) D4, CA = 54 years, 18–64-year model. Positive and negative SHAP values increase and decrease the model-estimated biological age, respectively.